\pdfoutput=1

\documentclass[11pt]{article}

\usepackage[]{acl}
\usepackage{url}

\usepackage{times}
\usepackage{latexsym}

\usepackage[T1]{fontenc}

\usepackage[utf8]{inputenc}

\usepackage{pifont} 

\usepackage{amsmath}
\usepackage{amsfonts}
\usepackage{lipsum} 
\usepackage{algorithm}
\usepackage{setspace}
\usepackage{graphicx}
\usepackage{tabularx}

\usepackage[disable=footnote]{microtype}
\usepackage{xcolor}
\definecolor{bluenode}{HTML}{00a7db}
\definecolor{rednode}{HTML}{ea4e00}
\definecolor{purplenode}{HTML}{9d00d7}
\definecolor{graynode}{HTML}{898989}
\definecolor{ForestGreen}{RGB}{34,139,34}
\usepackage{multirow}
\usepackage{makecell}
\usepackage{arydshln}
\usepackage{booktabs}

\usepackage{amsmath,amsfonts,bm}

\def\eqref#1{equation~\ref{#1}}

\def\1{\bm{1}}

\DeclareMathAlphabet{\mathsfit}{\encodingdefault}{\sfdefault}{m}{sl}
\SetMathAlphabet{\mathsfit}{bold}{\encodingdefault}{\sfdefault}{bx}{n}

\usepackage{mathtools}
\usepackage{xspace}
\newcommand{\eg}{\textit{e.g.}}

\newcommand{\methodname}{CoTasks\xspace}

\usepackage{ulem}
\usepackage{xcolor}
\DeclareRobustCommand{\erase}{\bgroup\markoverwith{\textcolor{red}{\rule[.5ex]{2pt}{0.4pt}}}\ULon}

\title{CoTasks: Chain-of-Thought based Video Instruction Tuning Tasks}

\author{
  Yanan Wang\thanks{~~~Equal contribution.} \quad
  Julio Vizcarra\footnotemark[1] \quad
  Zhi Li \quad
  Hao Niu \quad
  Mori Kurokawa \\
  KDDI Research, Inc. \\
  \texttt{\{wa-yanan, xju-vizcarra, zh-li, ha-niu, mo-kurokawa\}@kddi.com}
}

\begin{document}
\maketitle

\begin{abstract}
Despite recent progress in video large language models (\textbf{VideoLLMs}), a key open challenge remains: how to equip models with chain-of-thought (\textbf{CoT}) reasoning abilities grounded in fine-grained object-level video understanding. Existing instruction-tuned models, such as the Qwen and LLaVA series, are trained on high-level video-text pairs, often lacking structured annotations necessary for compositional, step-by-step reasoning. We propose \textbf{CoTasks: Chain-of-Thought based Video Instruction Tuning Tasks}, a new framework that decomposes complex video questions of existing datasets (\eg, NeXT-QA, STAR) into four entity-level foundational tasks: frame localization, entity tracking, spatial and temporal relation extraction. By embedding these intermediate CoT-style reasoning steps into the input, CoTasks enables models to explicitly perform object-centric spatiotemporal reasoning. Experiments on the NeXT-QA benchmark show that CoTasks significantly enhance inference performance: LLaVA-video-7B improves by \textbf{+3.3} points in average GPT-4 evaluation score, and Qwen2.5-VL-3B gains \textbf{+17.4}, with large boosts in causal (\textbf{+14.6}), temporal (\textbf{+10.9}) and descriptive (\textbf{+48.1}) subcategories. These results demonstrate the effectiveness of CoTasks as a structured CoT-style supervision framework for improving compositional video reasoning.
\end{abstract}

\section{Introduction}

\begin{figure}[ht]
    \centering
    \includegraphics[width=\linewidth]{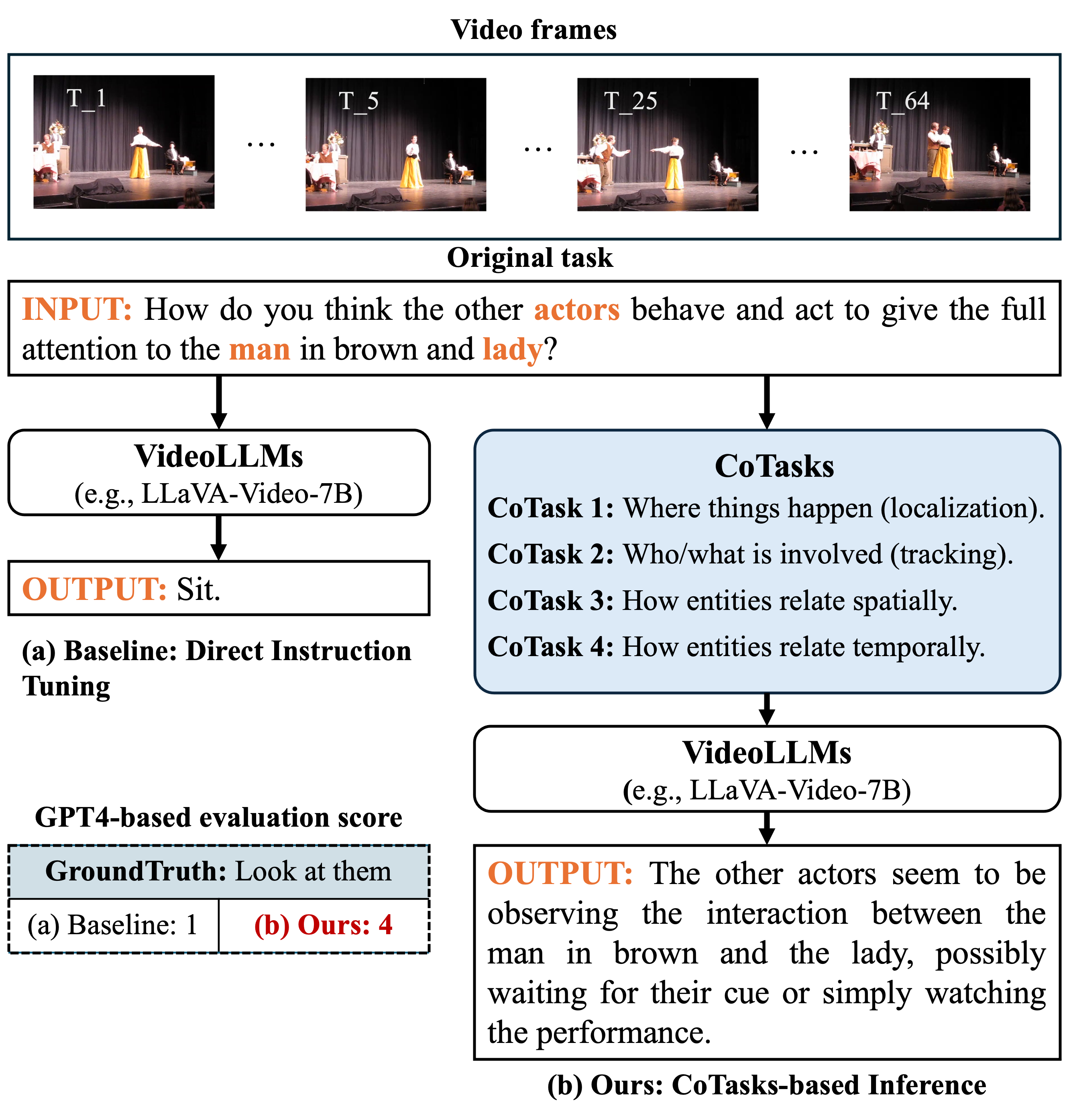}
    \caption{
    Comparison of LLaVA-video-7B with and without CoTasks. Given a complex video question, direct inference yields a shallow answer (``Sit''), whereas CoTasks guides the model through entity-aware reasoning, resulting in a more grounded and descriptive response.
    }
    \label{fig:fig1}
\end{figure}

Video Large Language Models (VideoLLMs) are rapidly gaining importance in applications such as \textit{interactive video QA systems}, including ChatGPT~\cite{openai2023gpt4v} and Gemini~\cite{google2023gemini}, and \textit{embodied agents}, including OK-Robot~\cite{liu2024okrobot} and DynaMem~\cite{liu2024dynamem}.
These models aim to understand complex, dynamic visual scenes and generate coherent answers or decisions from high-level natural language queries.
Despite recent advances, a key open challenge remains: how to equip these models with structured reasoning abilities that reflect the step-by-step understanding humans apply when interpreting real-world videos.

Current state-of-the-art VideoLLMs, including the Qwen~\citep{Qwen-VL,Qwen2-VL,Qwen2.5-VL} and LLaVA~\citep{liu2023llava,zhang2024llava} families, are predominantly trained using high-level video instruction tuning. 
These models perform reasonably well on surface-level tasks, but often fail in scenarios that require nuanced object-level and temporal reasoning.
As shown in \autoref{fig:fig1}, given a complex video-based question, models like LLaVA-video-7B tend to produce shallow answers such as ``Sit'' failing to ground the response in the visual context or reflect the full narrative.
We hypothesize that this limitation arises from the absence of \textit{intermediate supervision signals}, which guide the model to reason about \textbf{where}, \textbf{who/what}, and \textbf{how} entities interact over time.
Inspired by Chain-of-Thought (CoT) prompting~\citep{wei2022chain}, we argue that injecting structured, step-wise visual cues can bridge this gap.

To this end, we introduce \textbf{CoTasks}, a new framework for \textit{Chain-of-Thought based Video Instruction Tuning Tasks}. CoTasks decomposes high-level questions from existing video QA datasets (\eg, NeXT-QA~\citep{xiao2021next}, STAR~\citep{wu2021star}) into four foundational reasoning subtasks: (1) \textit{frame localization}, (2) \textit{entity tracking}, (3) \textit{spatial relation extraction}, and (4) \textit{temporal relation extraction}. 
By embedding these intermediate CoT-style reasoning steps into the input, CoTasks enables models to explicitly perform object-centric spatiotemporal reasoning.  
As a result, giving a high-level question, the model can ground and compose visual evidence step-by-step before producing an answer.
As visualized in \autoref{fig:fig1}, our approach transforms the vague output of standard VideoLLMs into a much richer response that closely mirrors human understanding.

We evaluate CoTasks using three VideoLLMs, \textit{Qwen2.5-VL-3B}, \textit{Qwen2.5-VL-7B}, and \textit{LLaVA-video-7B} on the NeXT-QA and STAR benchmarks.
To assess the upper bound of performance gains, we prompt models with the original question augmented by the ground-truth answers of the CoTasks subtasks. This allows us to isolate how much structured CoT-style context can contribute to final answer quality. 
Our inference-time augmentation strategy requires no architectural changes or re-training.
CoTasks significantly boosts performance across all models. Notably, Qwen2.5-VL-3B—a lightweight model—achieves a \textbf{+17.4} point gain in average GPT-4 evaluation score, with large improvements in \textit{causal} (\textbf{+14.6}), \textit{temporal} (\textbf{+10.9}), and \textit{descriptive} (\textbf{+48.1}) reasoning categories. These results demonstrate the effectiveness of CoTasks as a lightweight yet powerful mechanism for enhancing structured video reasoning.
Moreover, our findings highlight the promise of using CoTasks as an instruction-tuning curriculum for building high-performance, resource-efficient VideoLLMs—a key requirement for on-device, real-world applications.

\section{Problem setup}\label{sec:problem}
In this work, we focus on constructing a video instruction tuning dataset grounded in chain-of-thought (CoT) reasoning. Given object-centric video question answering (VideoQA) tasks such as NeXT-QA and STAR—which provide fine-grained annotations including object bounding boxes, intra-frame spatial relations, and inter-frame temporal relations—we reformulate the original multiple-choice questions into open-ended, free-form answering tasks. Furthermore, we augment each complex visual reasoning question with four foundational CoTasks: frame localization, object tracking, spatial relation extraction, and temporal relation extraction. The primary objective is to investigate whether these CoTasks can enhance the reasoning capabilities of state-of-the-art videoLLMs.



\section{Related work} \label{sec:related_work}

\noindent \textbf{Multimodal Large Language Models (MLLMs).} 
MLLMs, such as LLaVA~\cite{liu2023llava}, Qwen2-VL~\cite{Qwen-VL}, and LLaMA3-Vision~\cite{llama_3}, have extended language models into visual domains by integrating (1) a vision encoder~\cite{radford2021learning, tschannen2022siglip} to extract features from images or video frames, (2) a projection module to align visual features with the language embedding space, and (3) a language model backbone for multimodal reasoning. 
Recent works, including Video-ChatGPT~\cite{maaz2023videochatgpt}, LLaVA-Video~\cite{zhang2024llava}, Qwen2.5-VL~\cite{Qwen2.5-VL}, and AdaReTaKe~\cite{wang2025adaretake}, extend MLLMs to videos by compressing and aligning frame sequences to enable long-context understanding.
However, these models primarily focus on high-level instruction tuning without decomposing complex queries into structured sub-tasks---such as frame localization, object tracking, and spatiotemporal relation reasoning---limiting their ability to handle detailed object interactions or causal inference in dynamic scenes.
In contrast, our proposed \textbf{CoTasks} framework addresses this limitation by decomposing high-level video QA into four foundational sub-tasks: entity-based frame localization, object tracking, spatial relation extraction, and temporal relation extraction. 
These sub-tasks serve as intermediate, Chain-of-Thought (CoT)-style reasoning steps that inject structured guidance into video instruction tuning and enhance fine-grained spatiotemporal reasoning.

\noindent \textbf{Video Instruction Tuning dataset.}
Several large-scale video instruction datasets have recently been proposed to improve VideoLLMs' ability to follow complex visual-language instructions. LLaVA-Video-178K~\cite{zhang2024llava} is a synthetic dataset of 178K video samples, each annotated with detailed captions and open- and closed-ended questions. Generated using GPT-4 and human feedback, it supports instruction tuning for models like LLaVA-Video but focuses on high-level annotations without explicit intermediate reasoning structures. VideoInstruct-100K~\cite{maaz2023videochatgpt} consists of 100K video-instruction pairs with spatial, temporal, and reasoning-oriented questions, used to train Video-ChatGPT. While high-quality, it lacks structured object-centric decomposition, making it difficult to model fine-grained spatiotemporal reasoning. Video-STaR (VSTaR-1M)~\cite{zohar2025videostar} leverages diverse video sources (e.g., Kinetics-700~\cite{carreira2019short}, STAR~\cite{wu2021star}) to scale to 1M instruction pairs and aims to support general reasoning including CoT.
However, it relies on task-level supervision without sub-task grounding. 
In contrast, our proposed CoTasks framework explicitly decomposes complex video QA into four structured sub-tasks—frame localization, object tracking, spatial and temporal relation extraction—which are formatted as intermediate Chain-of-Thought (CoT) steps. 
This object-level supervision enhances compositional video reasoning and better aligns with human-like step-by-step understanding, addressing a key limitation in prior datasets.

\noindent\textbf{Chain-of-Thought Video Reasoning.}
Recent efforts have aimed to enhance VideoLLMs with step-by-step reasoning via Chain-of-Thought (CoT) prompting~\cite{wei2022chain}. LLaVA-o1~\cite{liu2024llavao1} introduces CoT-based vision-language instruction tuning, demonstrating improvements in visual question answering with structured step generation. Similarly, works such as Agent-of-Thoughts~\cite{chen2024agent}, Video-of-Thought~\cite{zhang2023video}, and Visual CoT~\cite{shao2024visual} explore CoT through modular reasoning agents, dynamic scene graphs, or benchmark-driven evaluation. SlowFocus~\cite{nie2024slowfocus} and MVU~\cite{chen2024mvu} incorporate fine-grained temporal and object-level cues to improve interpretability. 
However, most approaches depend on implicit supervision or architectural adaptations and lack systematic object-level decomposition aligned with CoT stages. 
In contrast, \textbf{CoTasks} proposes four interpretable and structured sub-tasks—frame localization, entity tracking, and spatial/temporal relation extraction—that act as intermediate reasoning steps for improving CoT-based video instruction tuning.

\begin{figure*}[t]
    \centering
    \includegraphics[width=\linewidth]{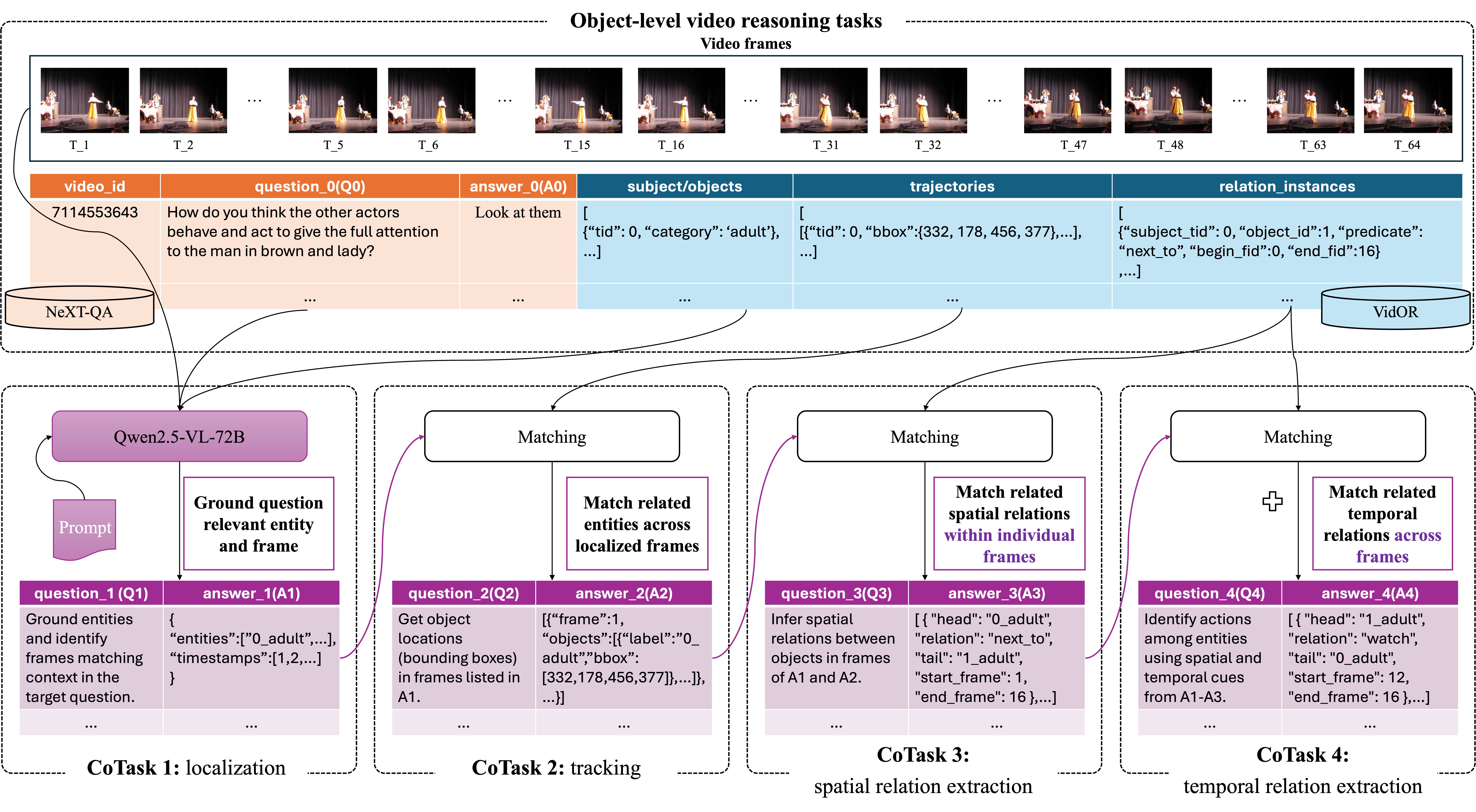}
    \caption{The figure illustrates the process of decomposing a VideoQA task (\eg, from NeXT-QA) into four structured CoTasks using object-level annotations (\eg, from VidOR). First, we reconstruct the VideoQA task to align video frames with object-level annotations (\S~\ref{reconstruction}). Then, we construct CoTasks based on the reconstructed VideoQA, where each CoTask builds upon the previous one (\S~\ref{cotasks-construction}).}
    \label{fig:fig2}
\end{figure*}

\section{Approach}

\subsection{CoTasks Definition} \label{CoTasks-definition}
Consider the original task defined as follows:

\begin{table}[h]
\centering
\small
\begin{tabularx}{0.95\linewidth}{X}
\toprule
\textbf{Original Task: Comprehensive Video Understanding} \\
\midrule
This task requires answering a high-level video question that demands object-level context and comprehensive scene understanding. \\

\vspace{0.1em}
\textbf{Question Format:} Open-ended question requiring holistic understanding of the video content. \\

\textbf{Answer Format:} Free-form short text. \\
\bottomrule
\end{tabularx}
\label{tab:originaltask_onecol}
\end{table}
We introduce a set of auxiliary tasks termed \textbf{CoTasks}. Each CoTask targets a distinct aspect of object-centric visual reasoning grounded in video understanding. We define the four foundational \textbf{\methodname} as follows:

\begin{table}[h]
\centering
\small
\begin{tabularx}{0.95\linewidth}{X}
\toprule
\textbf{CoTask 1: Object-based Frame Localization} \\
\midrule
This task involves identifying video frames where a specific combination of objects co-occur, as specified in the original question. \\
\vspace{0.1em}
\textbf{Question Format:} Ground entities and identify frames matching context in the target question.

\textbf{Answer Format:} A dictionary with the following keys:
\texttt{\{"objects": [str], "timestamps": [int]\}} \\

\bottomrule
\end{tabularx}
\label{tab:cotask1_onecol}
\end{table}

\begin{table}[h]
\centering
\small
\begin{tabularx}{0.95\linewidth}{X}
\toprule
\textbf{CoTask 2: Object Tracking with Bounding Boxes} \\
\midrule
This task requires tracking and localizing each mentioned object by generating bounding boxes over relevant video frames. \\

\vspace{0.1em}
\textbf{Question Format:} Get object locations (bounding boxes) in frames listed in CoTask 1.\\

\textbf{Answer Format:} A list of dictionaries, each with:
\texttt{[\{"frame": int, "objects": [\{"label": str, "bbox": [x1, y1, x2, y2]\}, ...]\}, ...]} \\
\bottomrule
\end{tabularx}
\label{tab:cotask2_onecol}
\end{table}

\begin{table}[h]
\centering
\small
\begin{tabularx}{0.95\linewidth}{X}
\toprule
\textbf{CoTask 3: Spatial Relation Extraction} \\
\midrule
This task involves extracting spatial relationships (\eg, \texttt{next\_to}, \texttt{in\_front\_of}) between pairs of objects within a frame span. \\

\vspace{0.1em}
\textbf{Question Format:} Infer spatial relations between objects in frames of CoTask 1 and CoTask 2. \\

\textbf{Answer Format:} A list of dictionaries with:
\texttt{[\{"head": str, "relation": str, "tail": str, "start\_frame": int, "end\_frame": int\}, ...]} \\
\bottomrule
\end{tabularx}
\label{tab:cotask3_onecol}
\end{table}

\begin{table}[h]
\centering
\small
\begin{tabularx}{0.95\linewidth}{X}
\toprule
\textbf{CoTask 4: Temporal Relation Extraction} \\
\midrule
This task identifies temporal interactions (\eg, \texttt{carry}, \texttt{follow}) between objects across a sequence of frames. \\

\vspace{0.1em}
\textbf{Question Format:} Identify actions among entities using spatial and temporal cues from CoTask 1-3. \\

\textbf{Answer Format:} A list of dictionaries with:
\texttt{[\{"head": str, "relation": str, "tail": str, "start\_frame": int, "end\_frame": int\}, ...]} \\
\bottomrule
\end{tabularx}
\label{tab:cotask4_onecol}
\end{table}

\subsection{CoTasks construction pipeline}
Figure~\ref{fig:fig2} illustrates the process of constructing CoTasks by decomposing a given video question answering task (\eg, from NeXT-QA) into four structured CoTasks using object-level annotations (\eg, VidOR~\cite{shang2019annotating}).

\subsubsection{Object-level video reasoning task reconstruction} \label{reconstruction}
We reconstruct the VideoQA task to align video frames with object-level annotations. In this work, we focus on two representative VideoQA benchmarks: NeXT-QA and STAR.
NeXT-QA is built upon the same raw video data as the VidOR dataset but lacks object-level annotations such as bounding boxes and spatial or temporal relations. To address this limitation, we merge the annotations from VidOR into NeXT-QA by matching \texttt{video\_id}s. We then uniformly sample $64$ frames per video, along with their corresponding annotations. The choice of $64$ frames is empirically validated to be optimal for enabling LLaVA-Video to function effectively (see Table~\ref{tab:gpt4_comparison} and Table~\ref{tab:cotasks_ablation}).
In contrast, STAR is natively constructed as an object-centric VideoQA dataset. It includes the full set of original video frames (ranging from $0$ to $92$) and provides comprehensive object-level annotations, which we directly leverage for CoTask construction.

\subsubsection{CoTasks construction} \label{cotasks-construction}
As shown in Figure~\ref{fig:fig2}, we construct four foundational CoTasks from the original object-level VideoQA tasks, following the definitions introduced in Section~\ref{CoTasks-definition}. In NeXT-QA, the high-level questions (\textit{question\_0 (Q0)}) are manually annotated by referring to the entire video rather than specific frames. As a result, we cannot directly ground the relevant video frames for \textit{Q0} without leveraging pretrained VideoLLMs. To address this, we utilize the state-of-the-art VideoLLM, Qwen2.5-VL-72B, to ground relevant objects and frames for \textit{Q0}, generating \textit{answer\_1 (A1)} for CoTask 1. The inputs for this step include the filtered video frames, \textit{Q0}, and identified subject/object entities, along with the prompt template shown in Table~\ref{tab:cotask1_gen_prompt}.
In contrast, the \textit{Q0} questions in STAR are constructed with reference to relevant frames and their corresponding object-level annotations. We extract the grounded objects and frames directly from the STAR dataset and convert them into the CoTask 1 format without requiring a VideoLLM.
CoTask 2 is built upon \textit{A1} and aims to generate \textit{question\_2 (Q2)} by matching objects and their bounding boxes across the localized frames. Its inputs are \textit{A1} and object trajectories.
CoTask 3 focuses on extracting spatial relations within individual localized frames. The inputs for this task include \textit{A1}, \textit{A2}, and spatial \textit{relation\_instances}.
Finally, CoTask 4 targets temporal relation extraction across frames, using \textit{A1}, \textit{A2}, \textit{A3}, and temporal \textit{relation\_instances} as input.

\begin{table}[t]
\centering
\small
\begin{tabularx}{\linewidth}{X}
\toprule
\textbf{\#\#\# System:} \\
You are a vision-language reasoning assistant. You will be shown a sequence of 64 video frames and a question referring to entities and their interactions. Your job is to match the context in the question to one or more of the ground-truth entities, and determine which frames show those entities co-occurring or interacting. \\

\vspace{0.05em}
\textbf{\#\#\# Task:} \\
- Extract a list of entities (from the set above) that are mentioned or implied in the question (must be 1 or more). \\
- Identify the frame indices (timestamps) between 1 and 64 where those entities appear together.\\
- Return a concise answer as a single JSON object without any markdown code fences or backticks. \\
- The number of timestamps must be greater than 0 and less than 17.\\
- Include only the most relevant keyframes that are clearly related to the question.\\

\vspace{0.05em}
\textbf{\#\#\# Output format:} \\
\texttt{\{"entities": [/* list of extracted entity names */], "timestamps": [/* list of frame numbers from 1 to 64, length 1–16 */]\}} \\

\vspace{0.05em}
\textbf{\#\#\# Example:} \\
Question: why did the adult point to the fruits on the table? \\
Ground-truth entities: \texttt{['0\_baby', '1\_table', '2\_fruits', '3\_fruits', '4\_adult']} \\

\vspace{0.05em}
\textbf{\#\#\# Answer:} \\
\texttt{\{"entities": ['0\_baby', '2\_fruits', '4\_adult'], "timestamps": [1, 2, 8, 10]\}} \\

\vspace{0.05em}
\textbf{\#\#\# Your Turn:} \\
Question: \{\{YOUR\_QUESTION\_HERE\}\} \\
Ground-truth entities: \{\{Ground-truth entities\}\} \\
Answer:
\\
\bottomrule
\end{tabularx}
\caption{Prompt template for CoTask 1 data generation: Frame localization based on object co-occurrence.}
\label{tab:cotask1_gen_prompt}
\end{table}
\section{Experiments}\label{sec:experiment}

In this section, we evaluate the proposed CoTasks (CoTasks-NeXT-QA and CoTasks-STAR) using recent VideoLLMs (\eg, Qwen2.5-VL-3/7/72B and LLaVA-Video-7B) to assess their impact on inference-time performance (\S\ref{sec:performance}). We further conduct an ablation study to examine the difficulty of solving CoTasks with these models, and to analyze the contribution of each CoTask component during inference (\S\ref{sec:ablation}). Specifically, we evaluate the contributions of CoTask 1-2 (frame localization and object tracking) for grounding, and CoTasks 3-4 (spatiotemporal relation extraction) for higher-level reasoning. In addition, we present qualitative visualizations of selected data samples to illustrate the quality of the constructed CoTasks (\S\ref{sec:case}). The generated responses are evaluated using a GPT-4-based automatic scoring framework~\cite{OpenEQA2023} (§\ref{sec:gpt4evaluation}).

\subsection{LLM-based Evaluator} \label{sec:gpt4evaluation}
We employ GPT-4 as an automatic evaluator to score the generated responses~\cite{OpenEQA2023}. Given the model outputs and corresponding ground-truth answers, along with the evaluation prompt template shown in Table~\ref{tab:gpt4_eval_prompt}, GPT-4 assigns a score from 1 to 5 based on their semantic alignment. The evaluation criteria are detailed in the prompt template.
\begin{table}[h]
\centering
\small
\begin{tabularx}{\linewidth}{X}
\toprule
\textbf{\#\#\# System:} \\
You are an AI assistant who will help me to evaluate the response given the question and the correct answer.
To mark a response, you should output a single integer between 1 and 5 (including 1, 5). \\
- 5 means that the response perfectly matches the answer. \\
- 1 means that the response is completely different from the answer. \\

\vspace{0.05em}
\textbf{\#\#\# Example 1:} \\
Question: Is it overcast? \\
Answer: no \\
Response: yes \\
Your mark: 1 \\

\vspace{0.05em}
\textbf{\#\#\# Example 2:} \\
Question: Who is standing at the table? \\
Answer: woman \\
Response: Jessica \\
Your mark: 3 \\

\vspace{0.05em}
\textbf{\#\#\# Example 3:} \\
Question: Are there drapes to the right of the bed? \\
Answer: yes \\
Response: yes \\
Your mark: 5 \\

\vspace{0.05em}
\textbf{\#\#\# Your Turn:} \\
Question: \{\{question\}\} \\
Answer: \{\{answer\}\} \\
Response: \{\{prediction\}\} \\
\bottomrule
\end{tabularx}
\caption{Prompt template for GPT-4-based evaluation of open-ended videoQA responses. Evaluators assign scores from 1–5 based on alignment with ground-truth and generated answers.}
\label{tab:gpt4_eval_prompt}
\end{table}

\subsection{Dataset} \label{sec:dataset}
\begin{figure}[h]
    \centering
    \includegraphics[width=0.9\linewidth]{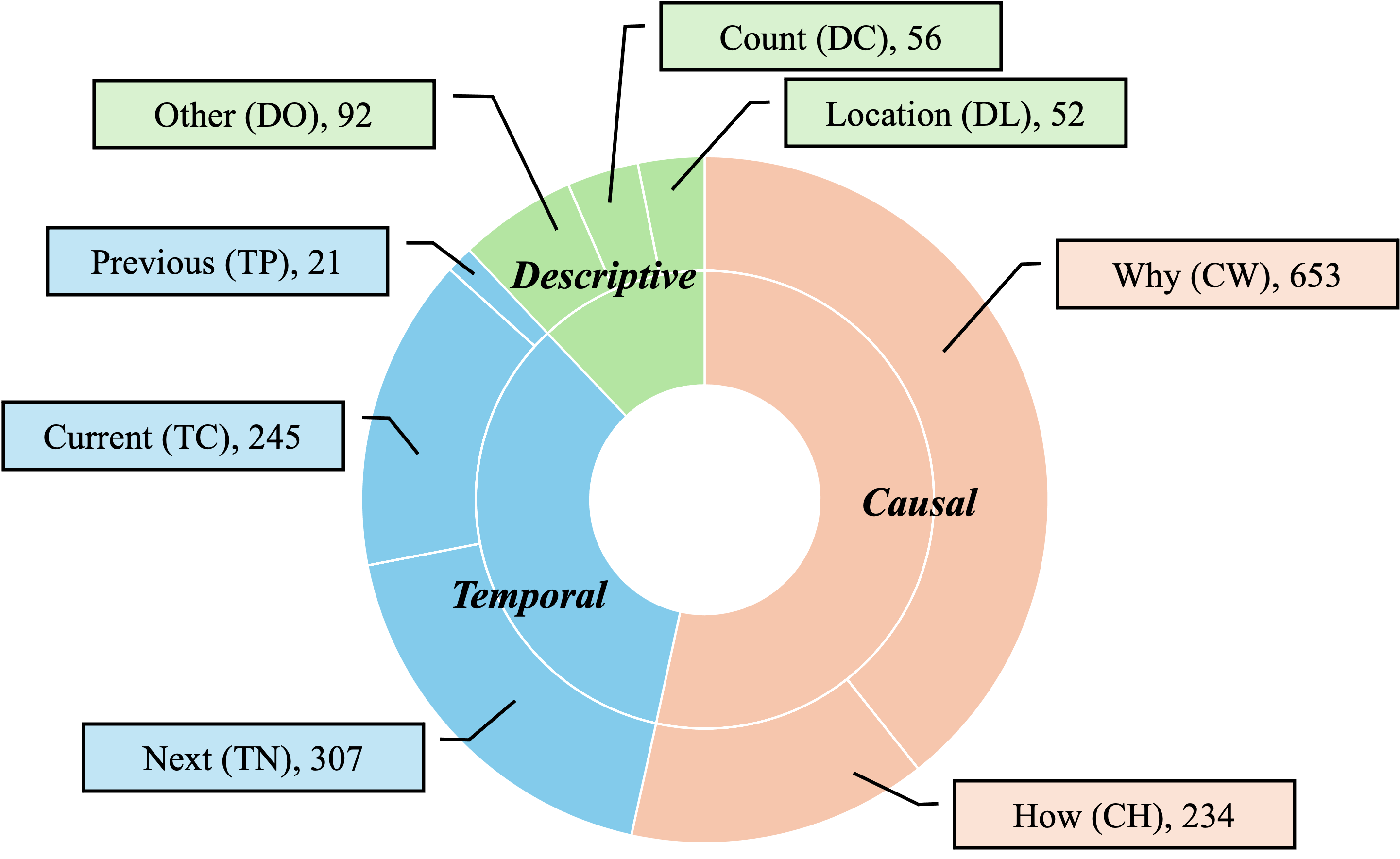}
    \caption{Distribution of question types in the CoTasks-NeXT-QA validation set. The inner ring groups questions into three high-level categories: Causal, Temporal, and Descriptive. The outer ring shows the corresponding fine-grained subcategories, illustrating the diversity of reasoning required.}
    \label{fig:val_question_donut}
\end{figure}

We construct two CoTask datasets based on existing object-centric VideoQA benchmarks: NeXT-QA and STAR, resulting in \textbf{CoTasks-NeXT-QA} and \textbf{CoTasks-STAR}, respectively.
For \textbf{CoTasks-NeXT-QA}, we utilize a subset of the original NeXT-QA samples due to limited access to the raw videos from the VidOR dataset. As shown in Table~\ref{tab:dataset_nextqa_stats}, the original dataset contains 5,440 videos and 47,692 questions. After filtering for available videos, we retain 3,821 videos and decompose each question into four foundational CoTask types, resulting in 43,392 CoTask samples. The distribution of question types in the validation set is illustrated in Figure~\ref{fig:val_question_donut}, with an outer ring showing subcategories and an inner ring grouping them into Causal, Temporal, and Descriptive types.
For \textbf{CoTasks-STAR}, we utilize all 3,946 videos available in the original STAR dataset. Each question is similarly decomposed into four CoTask questions, resulting in 211,316 CoTask samples, as summarized in Table~\ref{tab:dataset_star_stats}.

\begin{table}[h]
\centering
\small
\begin{tabular}{p{1.2cm}ccccc}
\toprule
\textbf{Dataset} & \textbf{Video} & \textbf{Train} & \textbf{Valid} & \textbf{Test} & \textbf{Total} \\
\midrule
Original & \textbf{5,440} & 34,132 & 4,996 & 8,564 & 47,692 \\
Filtered & 3,821 & 9,188 & 1,660 & - & 10,848 \\
\midrule
\textbf{CoTasks} & 3,821 & \textbf{36,752} & \textbf{6,640} & \textbf{-} & \textbf{43,392} \\
\bottomrule
\end{tabular}
\caption{Statistics for the CoTasks-NeXT-QA dataset. The “Original” row shows the number of questions in the original NeXT-QA dataset. The “Filtered” row reflects the subset for which video content is available. Each filtered question is reformulated into four foundational CoTask instances, resulting in the final statistics shown in the “CoTasks” row.}
\label{tab:dataset_nextqa_stats}
\end{table}

\begin{table}[h]
\centering
\small
\begin{tabular}{p{1.2cm}cccc}
\toprule
\textbf{Dataset} & \textbf{Video} & \textbf{Train} & \textbf{Valid} & \textbf{Total} \\
\midrule
Original & 3,946 & 45,731 & 7,098 & 52,829 \\
\midrule
\textbf{CoTasks} & 3,946 & 182,924 & 28,392 & 211,316 \\
\bottomrule
\end{tabular}
\caption{Statistics for the CoTasks-STAR dataset. The dataset includes all videos and questions from the original STAR benchmark. Each original question is reformulated into four CoTask questions, resulting in a substantial increase in total training and validation samples.}
\label{tab:dataset_star_stats}
\end{table}

\begin{table*}[h]
\centering
\small
\setlength{\tabcolsep}{4pt}
\begin{tabular}{lcccccccccc}
\toprule
\multirow{2}{*}{\textbf{Model}} & \multirow{2}{*}{\textbf{CoTasks}}
& \multicolumn{2}{c}{\textbf{Causal}} 
& \multicolumn{3}{c}{\textbf{Temporal}} 
& \multicolumn{3}{c}{\textbf{Descriptive}} 
& \multirow{2}{*}{\textbf{Avg}} \\
\cmidrule(lr){3-4} \cmidrule(lr){5-7} \cmidrule(lr){8-10}
& & CW & CH & TP & TC & TN & DC & DL & DO & \\
\midrule
\multirow{4}{*}{LLaVA-video-7B} 
& --       & 51.3 & 54.8 & 28.6 & 48.0 & 36.4 & 65.2 & 63.0 & 72.3 & 50.2 \\
& 1--2     & 53.4 & \textbf{55.9} & \textbf{32.1} & 49.4 & 38.0 & 70.5 & \textbf{72.1} & \textbf{78.5} & 52.6 \\
& 3--4     & 52.6 & 55.5 & \textbf{32.1} & 51.4 & 36.6 & \textbf{73.2} & 63.0 & 73.9 & 51.8 \\
& 1--4     & \textbf{55.1} & 54.8 & 27.4 & \textbf{51.9} & \textbf{39.3} & 70.5 & 68.8 & 76.9 & \textbf{53.5} \\
\bottomrule
\end{tabular}
\caption{
Ablation study on LLaVA-Video-7B across question types. CoTasks 1–2 (frame localization + tracking) and 3–4 (spatial/temporal relation extraction) yield distinct gains, while Combining all tasks (1--4) leads to the best average performance.
CW = Why, CH = How, TP = Previous, TC = Current, TN = Next, DC = Count, DL = Location, DO = Other.
}
\label{tab:cotasks_ablation}
\end{table*}

\subsection{Performance} \label{sec:performance}

We evaluate the effectiveness of CoTasks across state-of-the-art VideoLLMs using a GPT-4-based evaluator (see the prompt template in Table~\ref{tab:cotask5_eval_prompt} in the appendix). As shown in Table~\ref{tab:gpt4_comparison}, incorporating CoTasks consistently enhances model performance on CoTasks-NeXT-QA, with the most substantial improvement observed in the lightweight Qwen2.5-VL-3B model (+17.4 GPT-4 score). These results highlight a key limitation of current VideoLLMs in performing fine-grained, object-centric reasoning—particularly in smaller models. CoTasks help address this limitation by enriching inference with structured grounding and relational context.

Moreover, Table~\ref{tab:cotask_accuracy} shows that CoTasks significantly improve validation accuracy on the STAR dataset when using Qwen2.5-VL-3B, yielding a +34.3\% gain over the baseline without CoTask prompting. Notably, this performance also surpasses that of the model fine-tuned on the STAR training set by +13.8\%, demonstrating that structured grounding and relational context provided through CoTasks offer benefits beyond conventional instruction tuning with high-level question-answer pairs.

These findings suggest that CoTasks offer a promising approach for enhancing the reasoning capabilities of lightweight VideoLLMs, and they highlight a path toward developing fine-grained, object-centric VideoLLMs through structured multi-step prompting.

\begin{table}[h]
\centering
\small
\begin{tabular}{p{2.5cm}@{\hskip 3pt}cc@{\hskip 8pt}c@{\hskip 8pt}c@{\hskip 8pt}c}
\toprule
\textbf{Model} & \textbf{CT} & \textbf{C.} & \textbf{T.} & \textbf{D.} & \textbf{Avg.} \\
\midrule
\multirow{2}{=}{Qwen2.5-VL-3B} & -- & 35.0 & 21.6 & 13.8 & 27.8 \\
                               & \checkmark & \textbf{49.6} & \textbf{32.5} & \textbf{61.9} & \textbf{45.2} \\
\midrule
\multirow{2}{=}{Qwen2.5-VL-7B} & -- & 52.1 & 39.0 & 66.1 & 49.3 \\
                               & \checkmark & \textbf{55.3} & \textbf{39.8} & 66.1 & \textbf{51.3} \\
\midrule
\multirow{2}{=}{Qwen2.5-VL-72B} & -- & 52.7 & 38.9 & 67.6 & 49.7 \\
                                & \checkmark & \textbf{55.3} & \textbf{41.3} & \textbf{70.0} & \textbf{52.3} \\
\midrule
\multirow{2}{=}{LLaVA-video-7B} & -- & 52.2 & 41.1 & 67.9 & 50.2 \\
                                & \checkmark & \textbf{55.0} & \textbf{44.3} & \textbf{73.0} & \textbf{53.5} \\

\bottomrule
\end{tabular}
\caption{
GPT-4 evaluation scores with (\checkmark) and without (--) CoTasks-NeXT-QA. CT = CoTasks(1--4), C. = Causal, T. = Temporal, D. = Descriptive, Avg. = Average. 
\textbf{CoTasks consistently enhance performance across all models, yielding a notable +17.4 point gain for Qwen2.5-VL-3B.}
}
\label{tab:gpt4_comparison}
\end{table}



\begin{table}[h]
\centering
\small
\begin{tabular}{c c c c c}
\toprule
\textbf{Model} & \textbf{Dataset} & \textbf{PT} & \textbf{FT} & \textbf{Acc.(\%)} \\
\midrule
\multirow{3}{*}{Qwen2.5-VL-3B} & \multirow{3}{*}{STAR} & --         & --         & 31.1 \\
                               &                        & --         & \checkmark & 51.6 (+20.5) \\
                               &                        & \checkmark & --         & \textbf{65.4 (+34.3)} \\
\bottomrule
\end{tabular}
\caption{Accuracy on the STAR dataset using Qwen2.5-VL-3B under different combinations of CoTask-style prompting (PT) and fine-tuning (FT). Prompting alone significantly boosts performance, with gains of +34.3 points.}
\label{tab:cotask_accuracy}
\end{table}

\begin{figure*}[h]
    \centering
    \includegraphics[width=\linewidth]{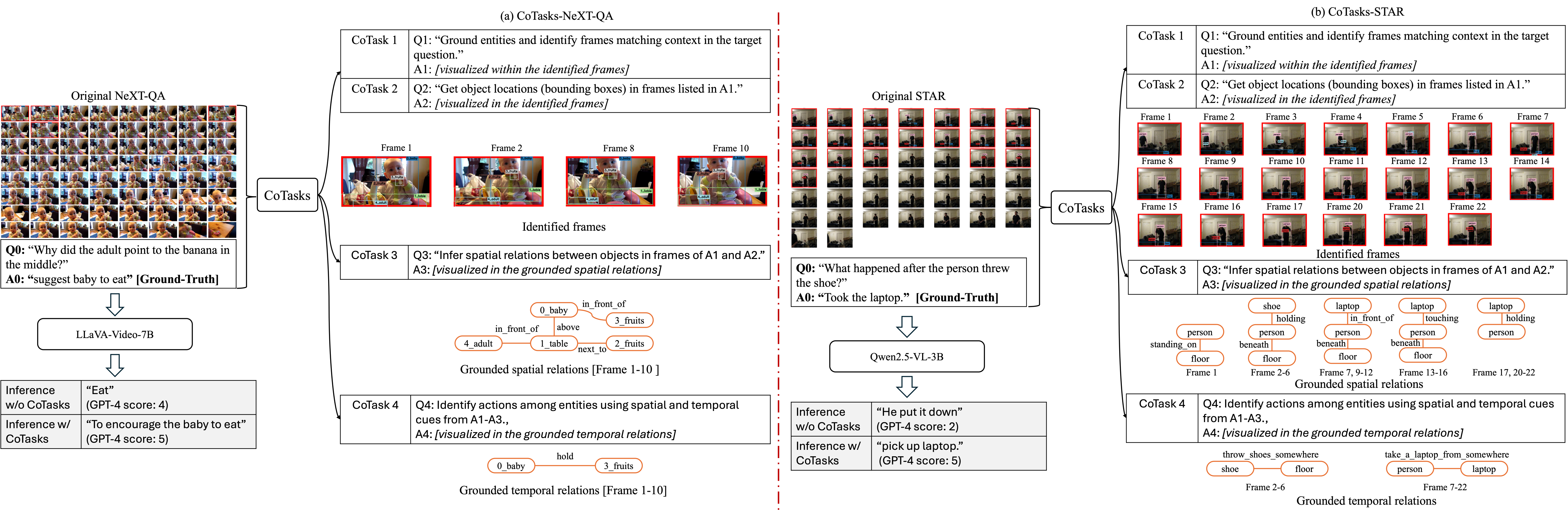}
    \caption{
Qualitative case study on CoTasks-NeXT-QA (left) and CoTasks-STAR (right) comparing model inference with and without CoTasks. Highlighted video frames are selected via CoTask 1 (frame localization), with object tracking (CoTask 2) shown via bounding boxes. Grounded spatial and temporal relations from CoTasks 3 and 4 are visualized below. CoTasks significantly improve response quality, as reflected in GPT-4 evaluation scores.
}
    \label{fig:fig4}
\end{figure*}
\subsection{Ablation study} \label{sec:ablation}

\paragraph{Per-CoTask Difficulty Analysis.} Table~\ref{tab:llava_cotask_val} examines the difficulty of solving CoTasks using the best-performing model identified in Table~\ref{tab:gpt4_comparison}. Specifically, we evaluate LLaVA-Video-7B on CoTasks 1–4 without fine-tuning, using the CoTasks-NeXT-QA validation set. All CoTasks yield GPT-4 evaluation scores below 30.0\%, revealing substantial limitations of current VideoLLMs in addressing fundamental visual reasoning tasks. These include CoTask 1 (frame localization), CoTask 2 (object tracking), CoTask 3 (spatial relation extraction), and CoTask 4 (temporal relation extraction). The results underscore the complementary nature of the CoTask design and highlight the need for targeted benchmarks to assess and improve foundational visual reasoning capabilities in VideoLLMs. The prompt templates used for this analysis are provided in Tables~\ref{tab:cotask1_eval_prompt}, \ref{tab:cotask2_eval_prompt}, \ref{tab:cotask3_eval_prompt}, \ref{tab:cotask4_eval_prompt} in the appendix.

\begin{table}[h]
\centering
\small
\setlength{\tabcolsep}{4pt}
\begin{tabularx}{\linewidth}{l *{4}{>{\centering\arraybackslash}p{1.35cm}}}
\toprule
\textbf{CoTask} & \textbf{C.} & \textbf{T.} & \textbf{D.} & \textbf{Avg.} \\
\midrule
CoTask 1 & 17.3 & 15.3 & 26.9 & 17.7 \\
CoTask 2 & 0.7  & 0.6  & 4.0  & 1.1  \\
CoTask 3 & 23.3 & 25.1 & 25.8 & 24.2 \\
CoTask 4 & 9.9  & 11.2 & 13.6 & 10.8 \\
\bottomrule
\end{tabularx}
\caption{Validation accuracy (\%) of \textbf{LLaVA-Video-7B} (64-frame input, no fine-tuning) on CoTasks 1–4, evaluated under GPT-4 rubric types: \textbf{C.} (causal), \textbf{T.} (temporal), and \textbf{D.} (descriptive).}
\label{tab:llava_cotask_val}
\end{table}

\paragraph{Impact of CoTask Subsets on Question Types.}
Table~\ref{tab:cotasks_ablation} presents an ablation study on LLaVA-Video-7B, evaluating the impact of different subsets of CoTasks on performance across various question types. Applying CoTasks 1–2 (frame localization and object tracking) improves average accuracy from 50.2 to 52.6, with notable gains in descriptive questions (e.g., +6.9 on DO and +7.3 on DL). CoTasks 3–4 (spatial and temporal relation extraction) also provide substantial improvements, especially for causal and descriptive categories, achieving the highest score on DC (count). When all CoTasks (1–4) are applied jointly, the model achieves the best overall performance (53.5 average), suggesting that each CoTask component contributes complementary information. 
These results highlight the future work on enhancing VideoLLMs' capabilities on structured, object-level reasoning tasks across diverse temporal and semantic categories.

\subsection{Case study} \label{sec:case}

Figure~\ref{fig:fig4} presents a qualitative case study comparing inference results with and without CoTasks on two examples—one from CoTasks-NeXT-QA using LLaVA-Video-7B (left) and one from CoTasks-STAR using Qwen2.5-VL-3B (right). In each case, the top rows show the full video sequence with the frames selected by CoTask 1 (frame localization) highlighted in red. Below, CoTask 2 visualizes object tracking using bounding boxes over localized frames. CoTask 3 and CoTask 4 illustrate spatial and temporal relationships between entities, grounded from the detected object instances.

The examples demonstrate that incorporating CoTasks significantly improves the quality of video understanding. For instance, the inference for the NeXT-QA sample is refined from a vague “Eat” (score: 4) to a more specific and context-aware response: “To encourage the baby to eat” (score: 5). Likewise, in the STAR sample, the model's prediction improves from “He put it down” (score: 2) to a precise action description: “Pick up laptop.” (score: 5). These results highlight how CoTasks contribute to grounding visual context and enhancing inference accuracy.

\section{Conclusions} \label{conclusions}

We present CoTasks, a structured framework for instruction tuning that enables VideoLLMs to perform chain-of-thought (CoT) reasoning grounded in fine-grained object-level video understanding. By decomposing complex video questions into four foundational sub-tasks—frame localization, entity tracking, spatial relation extraction, and temporal relation extraction—CoTasks provide explicit, interpretable supervision for spatiotemporal reasoning. Experiments on the NeXT-QA and STAR benchmarks demonstrate that CoTasks substantially improve inference performance, particularly for lightweight models such as Qwen2.5-VL-3B, which gains +17.4 GPT-4 points overall. The improvements span causal, temporal, and descriptive reasoning types, confirming the benefit of compositional task design. Our findings highlight the importance of structured, intermediate supervision for advancing compositional video understanding. Future work includes fine-tuning VideoLLMs directly on CoTasks and extending the framework to multi-modal and open-domain video reasoning settings.

\section*{Limitations}
Our approach presents several limitations. First, CoTasks require object-level annotations to expand existing VideoQA tasks, which may limit applicability in datasets lacking such annotations. Second, since each CoTask builds upon the preceding one, the quality of CoTask 1 (frame localization) directly affects the construction and reliability of subsequent CoTasks. Third, we evaluate model performance by prompting recent VideoLLMs with CoTasks rather than fine-tuning the models on the CoTask formulations. We leave fine-tuning with CoTasks as an important direction for future work.

\bibliography{custom}

\appendix
\section{Appendix}\label{sec:appendix}

\subsection{Answers to potential questions.}




\paragraph{Q1: Why were these 4 tasks selected for the CoTasks pipeline?}
A1: The combination of object-level tasks (object identification and tracking) with higher-level reasoning tasks (spatial relations and actions) creates a bridge for understanding fine-grained interactions in video, enabling more structured and compositional reasoning.

\paragraph{Q2: Is the pipeline for constructing CoTasks the same across different datasets (e.g., NeXT-QA, STAR)?}
A2: No, the pipeline differs for each dataset. For NeXT-QA, object-level information was retrieved from the VIDOR dataset annotations and aligned with the NeXT-QA questions and answers. In contrast, the STAR dataset is better structured and allows construction based on a combination of predefined catalogs, including objects, persons, relations, and object bounding boxes.

\paragraph{Q3: How is video selection and reduction to 64 frames performed?}
A3: A uniform sampling strategy is used to select 64 frames from the full set of video frames. These selected frames are then re-indexed to align with annotations, which originally refer to the full sequence of frames.

\paragraph{Q4: How do LLMs support the implementation of CoTasks?}
A4: Unlike exact textual matching, LLMs enable semantic understanding of questions and their referenced entities, allowing better alignment with ground-truth annotations for CoTasks construction.

\paragraph{Q5: Do you plan to open-source CoTasks?}
A5: Yes, we plan to open-source CoTasks after the review process.

\paragraph{Q6: Why don't you use multiple-choice style datasets?}
A6: Because the multiple-choice datasets provide all answer options as part of the input and require the model to select the correct one. This setup limits the model’s ability to generate free-form, compositional reasoning, which is essential for evaluating fine-grained understanding.


\subsection{Prompt template used in the evaluation.}
We provide all prompt templates used in Experiments (see \S\ref{sec:experiment}).

\begin{table}[h]
\centering
\small
\begin{tabularx}{\linewidth}{X}
\toprule
\textbf{\#\#\# Task:} \\
You are a vision-language reasoning model. Your goal is to extract relevant entities from a given question and identify the video frames (1–64) where those entities co-occur. \\

\vspace{0.05em}
\textbf{\#\#\# Instructions:} \\
- First, read the question carefully and determine which entities (from the grounded set) are directly mentioned or implied. \\
- Then, find the frames (from 1 to 64) where those entities appear together. \\
- Output your answer as a JSON object with two keys: \\
\quad - \texttt{"entities"}: list of relevant entity names (e.g., \texttt{"0\_adult"}, \texttt{"3\_handbag"}) \\
\quad - \texttt{"timestamps"}: list of frame numbers where those entities are present together \\
- The number of timestamps must be between 1 and 16. \\
- Do not include any extra explanation, markdown, or formatting—just return a valid JSON object. \\

\vspace{0.05em}
\textbf{\#\#\# Example Input:} \\
Q: Ground entities and identify frames matching context in the target question. \\
Contextual question: "What else does the man in yellow carry aside from a black laptop bag?" \\

\vspace{0.05em}
\textbf{\#\#\# Output format:} \\
\texttt{\{"entities": ["0\_adult", "3\_handbag"], "timestamps": [1, 5, 9, 12, 15]\}} \\

\vspace{0.05em}
\textbf{\#\#\# Your Turn:} \\
Q: Ground entities and identify frames matching context in the target question. \\
Contextual question: "What else does the man in yellow carry aside from a black laptop bag?" \\
Res:
\\
\bottomrule
\end{tabularx}
\caption{Prompt for CoTask 1: Entity grounding and timestamp prediction based on video QA context. The result shown in Table~\ref{tab:llava_cotask_val}.}
\label{tab:cotask1_eval_prompt}
\end{table}

\begin{table}[h]
\centering
\small
\begin{tabularx}{\linewidth}{X}
\toprule
\textbf{\#\#\# Task:} \\
You are a visual perception assistant. Based on a contextual question and prior grounding results, your task is to identify the bounding boxes of relevant entities in selected video frames. \\

\vspace{0.05em}
\textbf{\#\#\# Contextual question (Q0):} \\
What else does the man in yellow carry aside from a black laptop bag? \\

\vspace{0.05em}
\textbf{\#\#\# Reasoning question (Q2):} \\
Get object locations (bounding boxes) in frames listed in A1. \\

\vspace{0.05em}
\textbf{\#\#\# Grounded input (A1):} \\
\texttt{\{"entities": ["0\_adult", "3\_handbag"], "timestamps": [1, 5, 9, 12, 15]\}} \\

\vspace{0.05em}
\textbf{\#\#\# Instructions:} \\
- For each frame listed in \texttt{"timestamps"}, detect the presence of the listed \texttt{"entities"}. \\
- For each detected entity in a frame, return: \\
\quad - \texttt{"label"}: the entity ID (e.g., \texttt{"0\_adult"}, \texttt{"3\_handbag"}) \\
\quad - \texttt{"bbox"}: bounding box in the format \texttt{[x1, y1, x2, y2]} \\

- Output your result as a JSON list of dictionaries, each with: \\
\quad - \texttt{"frame"}: the frame number \\
\quad - \texttt{"objects"}: list of detected objects and their bounding boxes \\
- Do not include explanations, markdown, or extra text—only return valid JSON. \\

\vspace{0.05em}
\textbf{\#\#\# Output format example:} \\

\texttt{[\{"frame": 1, "objects": [{"label": "0\_adult", "bbox": [262, 2, 400, 333]}, {"label": "3\_handbag", "bbox": [294, 48, 393, 146]}]\},...]} \\

\vspace{0.05em}
\textbf{\#\#\# Your Turn:} \\
Contextual question: What else does the man in yellow carry aside from a black laptop bag? \\
Reasoning question: Get object locations (bounding boxes) in frames listed in A1. \\
Entities: \texttt{["0\_adult", "3\_handbag"]} \\
Frames: \texttt{[1, 5, 9, 12, 15]} \\
Res:
\\
\bottomrule
\end{tabularx}
\caption{Prompt for CoTask 2: Predicting bounding boxes for grounded entities across localized frames. The result shown in Table~\ref{tab:llava_cotask_val}.}
\label{tab:cotask2_eval_prompt}
\end{table}

\begin{table}[h]
\centering
\small
\begin{tabularx}{\linewidth}{X}
\toprule
\textbf{\#\#\# Task:} \\
You are a spatial reasoning assistant. Based on the contextual question and prior grounding information, your task is to infer spatial relationships between visual entities in specific video frames. \\

\vspace{0.05em}
\textbf{\#\#\# Contextual question (Q0):} \\
What else does the man in yellow carry aside from a black laptop bag? \\

\vspace{0.05em}
\textbf{\#\#\# Reasoning question (Q3):} \\
Infer spatial relations between objects in frames of A1 and A2. \\

\vspace{0.05em}
\textbf{\#\#\# Supporting input:} \\
\textbf{A1 (Entities and timestamps):} \\
\texttt{\{"entities": ["0\_adult", "3\_handbag"], "timestamps": [1, 5, 9, 12, 15]\}} \\

\textbf{A2 (Object bounding boxes per frame):} \\

\texttt{[\{"frame": 1, "objects": [\{"label": "0\_adult", "bbox": [262, 2, 400, 333]\}, \{"label": "3\_handbag", "bbox": [294, 48, 393, 146]\}]\}, \{"frame": 5, "objects": [\{"label": "0\_adult", "bbox": [355, 17, 520, 273]\}, \{"label": "3\_handbag", "bbox": [386, 0, 495, 87]\}]\}, \{"frame": 9, "objects": [\{"label": "0\_adult", "bbox": [369, 12, 480, 188]\}]\}, \{"frame": 12, "objects": [\{"label": "0\_adult", "bbox": [331, 14, 421, 140]\}]\}, \{"frame": 15, "objects": []\}]\}} \\

\vspace{0.05em}
\textbf{\#\#\# Instructions:} \\
- For each frame, determine if two entities are spatially related (e.g., \texttt{"next\_to"}, \texttt{"behind"}, \texttt{"on"}, etc.). \\
- A valid spatial relation must occur in individual frames. \\
- For each detected spatial relationship, return: \\
\quad - \texttt{"head"}: the source entity \\
\quad - \texttt{"relation"}: the spatial relationship \\
\quad - \texttt{"tail"}: the target entity \\
\quad - \texttt{"start\_frame"}: the first frame where the relation is observed \\
\quad - \texttt{"end\_frame"}: the last frame where the relation holds \\
- Output your result as a JSON list of dictionaries. \\
- Do not include explanations, markdown, or extra text—only return valid JSON. \\

\vspace{0.05em}
\textbf{\#\#\# Output format example:} \\
\texttt{[ \{"head": "0\_adult", "relation": "next\_to", "tail": "3\_handbag", "start\_frame": 1, "end\_frame": 5\}, ... ]} \\

\vspace{0.05em}
\textbf{\#\#\# Your Turn:} \\
Contextual question: What else does the man in yellow carry aside from a black laptop bag? \\
Reasoning question: Infer spatial relations between objects in frames of A1 and A2. \\
Entities: \texttt{["0\_adult", "3\_handbag"]} \\
Frames: \texttt{[1, 5, 9, 12, 15]} \\
Bounding boxes: see A2 above \\
Res:
\\
\bottomrule
\end{tabularx}
\caption{Prompt for CoTask 3: Inferring spatial relationships between grounded objects across localized frames. The result shown in Table~\ref{tab:llava_cotask_val}.}
\label{tab:cotask3_eval_prompt}
\end{table}

\begin{table}[h]
\centering
\small
\begin{tabularx}{\linewidth}{X}
\toprule
\textbf{\#\#\# Task:} \\
You are a visual reasoning agent. Your job is to analyze spatial and temporal cues from a sequence of video frames to infer action relationships between entities. \\

\vspace{0.05em}
\textbf{\#\#\# Contextual question (Q0):} \\
What else does the man in yellow carry aside from a black laptop bag? \\

\vspace{0.05em}
\textbf{\#\#\# Reasoning question (Q4):} \\
Identify actions among entities using spatial and temporal cues from A1--A3. \\

\vspace{0.05em}
\textbf{\#\#\# Supporting input:} \\

\textbf{A1 (Entities and timestamps):} \\
\texttt{\{"entities": ["0\_adult", "3\_handbag"], "timestamps": [1, 5, 9, 12, 15]\}} \\

\textbf{A2 (Bounding boxes per frame):} \\
\texttt{[\{"frame": 1, "objects": [\{"label": "0\_adult", "bbox": [262, 2, 400, 333]\}, \{"label": "3\_handbag", "bbox": [294, 48, 393, 146]\}]\}, \{"frame": 5, "objects": [\{"label": "0\_adult", "bbox": [355, 17, 520, 273]\}, \{"label": "3\_handbag", "bbox": [386, 0, 495, 87]\}]\}, \{"frame": 9, "objects": [...]\}, \{"frame": 12, "objects": [...]\}, \{"frame": 15, "objects": []\}]} \\

\textbf{A3 (Spatial relations):} \\
\texttt{[\{"head": "0\_adult", "relation": "next\_to", "tail": "3\_handbag", "start\_frame": 1, "end\_frame": 12\}]} \\

\vspace{0.05em}
\textbf{\#\#\# Instructions:} \\
- Infer \textbf{actions} between entities (e.g., \texttt{"carry"}, \texttt{"hold"}, \texttt{"push"}, \texttt{"pull"}) using: \\
\quad - Proximity and overlap in bounding boxes (A2) \\
\quad - Persistent spatial relations (A3) \\
- For each inferred action, return a dictionary with: \\
\quad - \texttt{"head"}: the acting entity \\
\quad - \texttt{"relation"}: the action verb \\
\quad - \texttt{"tail"}: the affected entity \\
\quad - \texttt{"start\_frame"}: first frame of the action \\
\quad - \texttt{"end\_frame"}: last frame of the action \\
- Return a list of such dictionaries in valid JSON format. \\
- Do not include explanations, markdown, or commentary—only the JSON. \\

\vspace{0.05em}
\textbf{\#\#\# Output format example:} \\
\texttt{[\{"head": "0\_adult", "relation": "carry", "tail": "3\_handbag", "start\_frame": 1, "end\_frame": 12\}]} \\

\vspace{0.05em}
\textbf{\#\#\# Your Turn:} \\
Contextual question: What else does the man in yellow carry aside from a black laptop bag? \\
Reasoning question: Identify actions among entities using spatial and temporal cues from A1--A3. \\
Entities: \texttt{["0\_adult", "3\_handbag"]} \\
Frames: \texttt{[1, 5, 9, 12, 15]} \\
Bounding boxes and spatial relations: see A2 and A3 above \\
Res:
\\
\bottomrule
\end{tabularx}
\caption{Prompt for CoTask 4: Temporal action reasoning based on bounding box and spatial relation history. The result shown in Table~\ref{tab:llava_cotask_val}.}
\label{tab:cotask4_eval_prompt}
\end{table}

\begin{table}[h]
\centering
\small
\begin{tabularx}{\linewidth}{X}
\toprule
\textbf{\#\#\# Task:} \\
You are a visual reasoning assistant. Given a series of video frames and visual annotations, your goal is to answer a high-level question (Q0) about an event involving specific entities. You will be provided supporting information from auxiliary visual sub-tasks (Q1--Q4), which ground entities, detect object locations, infer spatial relationships, and determine actions. \\

\vspace{0.05em}
Your response must be a concise phrase that best answers Q0 using reasoning based on the visual and relational evidence provided. \\

\vspace{0.05em}
\textbf{\#\#\# Instructions:} \\
- Use entity co-occurrence (A1), object bounding boxes (A2), spatial relations (A3), and inferred actions (A4) to support your answer. \\
- Base your answer on what the visual evidence consistently supports across the relevant frames. \\
- Respond with a \textbf{short phrase} (e.g., an object or action) directly answering Q0. \\

\vspace{0.05em}
\textbf{\#\#\# Input:} \\
Q0: what else does the man in yellow carry aside from a black laptop bag? \\
A1: \texttt{\{'entities': ['0\_adult', '3\_handbag'], 'timestamps': [1, 5, 9, 12, 15]\}} \\
A2: \texttt{[\{'frame': 1, 'objects': [\{'label': '0\_adult', 'bbox': [262, 2, 400, 333]\}, \{'label': '3\_handbag', 'bbox': [294, 48, 393, 146]\}\}, ...\}]} \\
A3: \texttt{[\{'head': '0\_adult', 'relation': 'next\_to', 'tail': '4\_handbag', 'start\_frame': 1, 'end\_frame': 12\}, ...]} \\
A4: \texttt{[\{'head': '0\_adult', 'relation': 'carry', 'tail': '4\_handbag', 'start\_frame': 1, 'end\_frame': 12\}, ...]} \\

\vspace{0.05em}
\textbf{\#\#\# Output format:} \\
Respond with a short phrase that answers Q0 using the evidence from A1--A4. \\

\textbf{Respond:} book \\
\bottomrule
\end{tabularx}
\caption{Prompt template for evaluating the original task (the result shown in Table~\ref{tab:gpt4_comparison}).}
\label{tab:cotask5_eval_prompt}
\end{table}

\end{document}